\makeatletter\providecommand{\@LN}[2]{}\providecommand{\@LN@col}[1]{}\makeatother
\RequirePackage{fix-cm}
\PassOptionsToPackage{table}{xcolor}
\documentclass[11pt,letterpaper,shortlabels]{berkeley}
\usepackage[authoryear,round]{natbib}
\usepackage{mathtools}
\usepackage{multirow,makecell,array,tabularx}
\usepackage{xspace}
\usepackage{wrapfig}
\usepackage{cuted}
\usepackage{currfile}
\usepackage[capitalize]{cleveref}
\crefname{section}{Sec.}{Secs.}
\Crefname{section}{Section}{Sections}
\Crefname{table}{Table}{Tables}
\crefname{table}{Tab.}{Tabs.}

\definecolor{NDblue}{RGB}{23,50,77}
\definecolor{NDgold}{RGB}{197,139,34}
\definecolor{cvprblue}{rgb}{0.21,0.49,0.74}
\definecolor{pscol}{HTML}{B3261E}
\definecolor{sfcol}{HTML}{1F5F8B}
\definecolor{darkgreen}{HTML}{006E33}
\hypersetup{colorlinks=true,linkcolor=NDblue,citecolor=NDgold,urlcolor=NDblue,
  pdftitle={Phantom Success, Silent Failure: Auditing the Success Checkers of Simulated Manipulation Benchmarks},
  pdfauthor={Zirui Song, Jiadong Zhang, Zheng Lu, Xiaosong Ma, Xiaojun Chang, Xiuying Chen}}
\renewcommand{\titlefont}{\color{BerkeleyBlue}\normalfont\bfseries
  \fontsize{17.5}{20}\selectfont}
\let\cite\citep
\newcommand{\tbd}[1]{{\color{red}[TBD: #1]}}
\newcommand{\ps}{phantom success}
\newcommand{\sfail}{silent failure}
\newcommand{\Ps}{Phantom success}
\newcommand{\Sfail}{Silent failure}
\newcommand{\simaudit}{SimAudit}

\newcommand{\xm}[1]{}            %
\makeatletter
\DeclareRobustCommand\onedot{\futurelet\@let@token\@onedot}
\def\@onedot{\ifx\@let@token.\else.\null\fi\xspace}

\def\etal{\emph{et al}\onedot}
\makeatother

\title{Phantom Success, Silent Failure: Auditing the Success Checkers of Simulated Manipulation Benchmarks}
\newcommand{\paperinstitutionsymbol}[1]{%
  \textcolor{NDblue}{\ensuremath{\ifcase#1\relax
    \or\spadesuit\or\heartsuit\or\clubsuit\or\diamondsuit\fi}}}
\newcommand{\papercontactmark}{\textcolor{NDgold}{*}}
\newcommand{\paperauthor}[2]{%
  \mbox{#2\textsuperscript{\paperinstitutionsymbol{#1}}}}
\newcommand{\paperaffil}[2]{%
  \mbox{\textsuperscript{\paperinstitutionsymbol{#1}}\,#2}}
\newcommand{\papertitlerule}{%
  {\color{NDgold}\rule{26pt}{0.6pt}}%
  {\color{NDblue!28}\rule{\dimexpr\linewidth-26pt\relax}{0.6pt}}}
\makeatletter
\renewcommand{\@author}{%
  {\normalfont\fontsize{11}{14}\selectfont
   \paperauthor{1}{Zirui~Song}\hspace{0.7em}%
   \paperauthor{1}{Jiadong~Zhang}\hspace{0.7em}%
   \paperauthor{2}{Zheng~Lu}\hspace{0.7em}%
   \paperauthor{1}{Xiaosong~Ma}\hspace{0.7em}%
   \paperauthor{1}{Xiaojun~Chang}\hspace{0.7em}%
   \paperauthor{1}{Xiuying~Chen}\textsuperscript{\papercontactmark}\par}
  \vspace{6pt}%
  {\normalfont\fontsize{10.5}{14}\selectfont
   \paperaffil{1}{Mohamed bin Zayed University of Artificial Intelligence}\hspace{1.2em}%
   \paperaffil{2}{Peking University}\par}}
\renewcommand{\maketitle}{%
  \begingroup
  \setlength{\parindent}{0pt}%
  \setlength{\parskip}{0pt}%
  \centering
  \vspace*{5pt}%
  {\titlefont\@title\par}
  \vspace{11pt}%
  \@author
  \vspace{8pt}%
  \papertitlerule\par
  \vspace{10pt}%
  \endgroup
  {\abscontent}%
  \thispagestyle{firststyle}}
\makeatother
\fancypagestyle{firststyle}{%
  \fancyhf{}%
  \fancyhead[L]{\raisebox{-0.3\height}{\includegraphics[height=44pt]{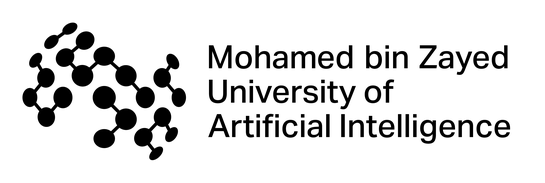}}%
    \hspace{16pt}%
    \raisebox{-0.3\height}{\includegraphics[height=44pt]{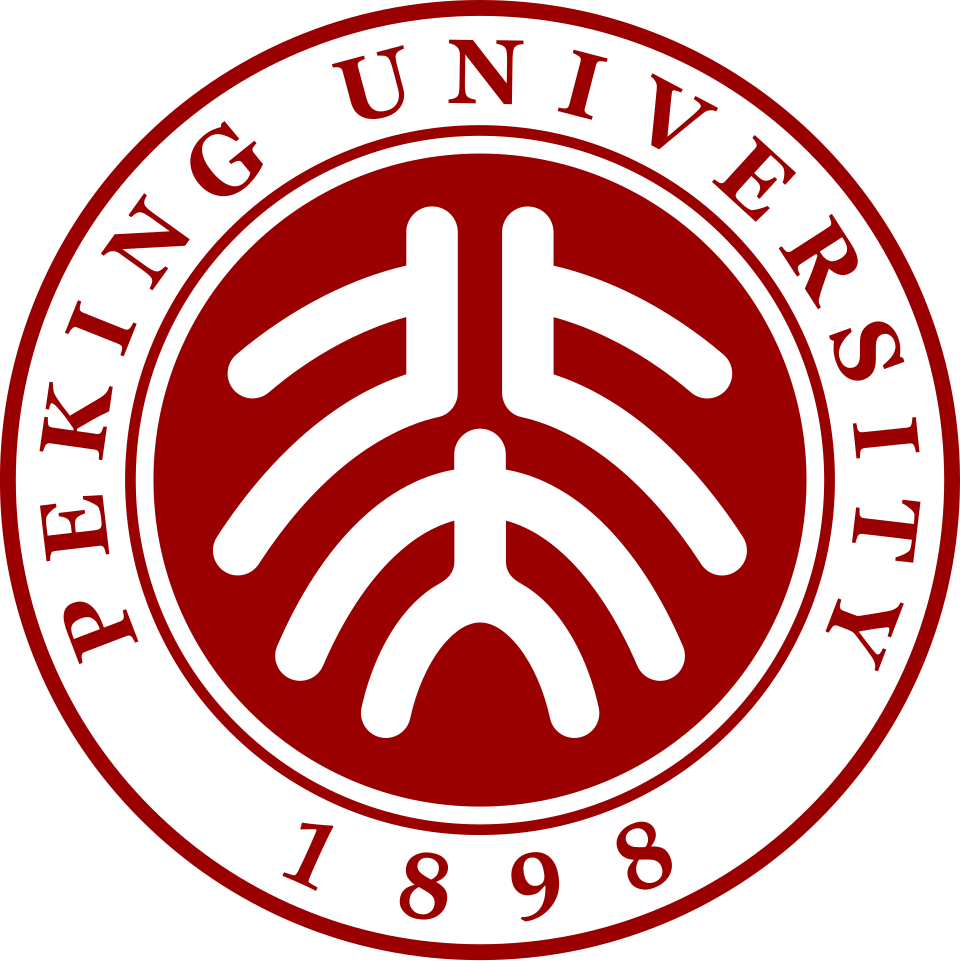}}}%
  \fancyfoot[L]{\normalfont\fontsize{9}{11}\selectfont
    \textsuperscript{\papercontactmark}\,Corresponding author:
    \the\correspondingauthor}}

\correspondingauthor{Xiuying~Chen (\href{mailto:xiuying.chen@mbzuai.ac.ae}{xiuying.chen@mbzuai.ac.ae})}
\begin{abstract}
Simulated manipulation benchmarks measure policy performance using hand-written success checkers that label each episode as a success or failure. 
Passing an expert demonstration shows that a checker accepts one way of completing the task. 
Learned policies, however, may stop after partial progress or complete the task in a different way.
A checker may then accept an incomplete execution (phantom success) or reject an execution that satisfies the instruction (silent failure).
In this paper, we introduce SimAudit, an LLM-based pipeline that audits success checkers against task instructions. 
SimAudit analyzes the instruction and checker code to identify scenarios where the two may disagree, recreates those scenarios in simulation, and runs the original checker. 
Multi-view renderings are then used to verify whether the observed behavior satisfies the instruction. 
Across RoboTwin 2.0, RoboCasa365, and RoboDojo, 183 simulated test cases produced the predicted checker verdict. 
Visual verification and deduplication yielded 61 distinct defects across 49 tasks, including 40 phantom successes and 21 silent failures. 
Most defects in RoboCasa365 accepted incomplete executions, most in RoboDojo rejected valid executions, and both error types were common in RoboTwin 2.0. 
These findings suggest that validating success checkers requires testing both incomplete executions and valid alternatives to expert demonstrations, and motivate more rigorous evaluation protocols for future benchmarks.

\end{abstract}
\begin{document}
\raggedbottom
\maketitle
\section{Introduction}
\label{sec:intro}

Task success rate is the primary metric for comparing robot manipulation policies in simulation.
Each episode is scored by a \emph{success checker}, a hand-written rule that uses simulator state to determine whether the robot completed the task specified by the instruction. 
Evaluating generalist manipulation models, including vision-language-action models~\cite{brohan2023rt2visionlanguageactionmodelstransfer,kim2024openvlaopensourcevisionlanguageactionmodel,black2026pi0visionlanguageactionflowmodel,DBLP:conf/aaai/SongOLJWXZZJJCL26,lu2026layerroute} and world action models~\cite{li2025unifiedvideoactionmodel,bi2025motusunifiedlatentaction,lin2026rifar}, on physical robots is costly and difficult to reproduce across laboratories~\cite{DBLP:journals/corr/abs-2503-24278,202606.0425}.
Recent simulated benchmarks such as RoboTwin~2.0~\cite{DBLP:journals/corr/abs-2506-18088}, RoboCasa365~\cite{DBLP:conf/iclr/NasirianyNMZ26}, and RoboDojo~\cite{chen2026robodojounifiedsimandrealbenchmark} therefore play a central role in model evaluation. 
They aggregate checker verdicts into the success rates used to rank policies.
Prior work has shown that these rates can be affected by memorization of training configurations~\cite{zhou2026liberoprorobustfairevaluation,fei2025liberoplusindepthrobustnessanalysis}, task shortcuts and overfitting to fixed test sets~\cite{jiang2026actuallybenchmarkingrobotmanipulation}, and artifacts in the physics engine~\cite{zhou2026simulatorfailspolicyauditing}.
The reliability of the success checkers that produce these rates has
received less attention. 
This raises a basic question: do checker verdicts
accurately reflect whether a robot's execution satisfies the task instruction?

\begin{figure}[t]
  \centering
  \includegraphics[width=\linewidth]{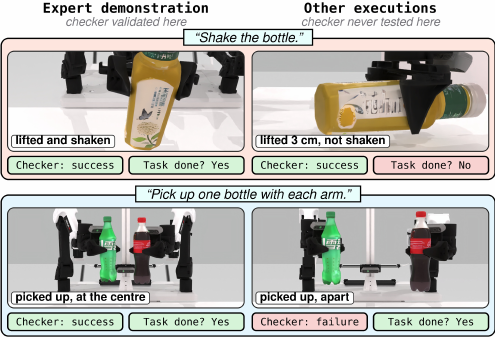}
 \caption{\textbf{A checker validated on the expert can be wrong elsewhere.}
On the expert demonstration (left), the checker agrees with the instruction.
On other executions (right), the same checker accepts a bottle that is lifted
but not shaken, and rejects two bottles that are picked up away from the centre.}
  \label{fig:concept}
\end{figure}

Expert demonstrations can establish that a success checker accepts a reference execution.
RoboTwin~2.0~\cite{DBLP:journals/corr/abs-2506-18088}, for example, retains an evaluation seed only if the scripted expert passes the checker on that seed. 
A learned policy can reach states outside those covered by the demonstrations by stopping before the task is complete or by completing it in a different way. 
Figure~\ref{fig:concept} illustrates two errors that can arise in these states. 
The first task illustrates a \emph{\ps{}}. 
The checker reports success when the robot lifts the bottle 3\,cm off the table without shaking it.
A \ps{} occurs when the checker marks an execution as successful even though it does not satisfy the instruction.
The second task illustrates a \emph{\sfail{}}. 
The robot picks up one bottle with each arm as instructed, but the checker reports failure because it additionally requires both bottles to be at the centre of the table.
A \sfail{} occurs when the checker marks an execution as failed even though it satisfies the instruction.

Validation with scripted experts can miss checker errors because expert trajectories may never expose discrepancies between the instruction and the conditions the checker tests. 
To uncover these errors, we introduce \simaudit{}, an LLM-based pipeline that audits success checkers on states that scripted experts never reach.
Given a task instruction and its checker code, an LLM agent compares the requirements in the instruction with the conditions tested by the checker. It uses discrepancies between the two to propose states that the checker may misclassify. 
For each proposed state, the agent predicts from the code whether the checker will return success or failure. The agent then constructs the state in simulation, using robot motion governed by simulated physics wherever possible. The pipeline runs the unmodified checker on the resulting state and renders the scene from multiple camera views. A candidate is retained only if the rendered views establish that the execution's outcome under the instruction differs from the checker verdict.
Across 142 tasks in the three benchmarks, the agent proposed 568 candidate states.
The actual checker verdict matched the agent's prediction in 183 cases. 
Visual verification and deduplication yielded 61 distinct checker defects.

We identify 61 defects across 49 tasks in RoboTwin~2.0, RoboCasa365, and RoboDojo, including 40 \ps{}es and 21 \sfail{}s. RoboCasa365 has 17 \ps{}es and one \sfail{}, while RoboDojo has three \ps{}es and nine \sfail{}s. Both types are common in RoboTwin~2.0, with 20 \ps{}es and 11 \sfail{}s.
These patterns reflect the conditions each benchmark's checkers enforce. RoboCasa365 checkers often omit conditions required by the instruction, allowing incomplete executions to pass. RoboDojo checkers often impose conditions beyond those in the instruction, causing executions that satisfy the instruction to fail.
These errors can inflate or depress reported success rates. To measure their effect on released policies, we repair the defects identified by \simaudit{} and score the same policy rollouts with the original and repaired checkers. We count a verdict change whenever the two checkers disagree on an episode.

Our contributions are threefold. First, we define two types of success checker errors, \ps{} and \sfail{}, and illustrate both with rendered examples from states outside expert trajectories. Second, we introduce \simaudit{}, an agentic pipeline that proposes candidate error cases from task instructions and checker code, constructs them in simulation, and verifies errors by comparing the unmodified checker's verdict with renders from multiple camera views. Third, we report 61 visually verified defects across three benchmarks, with different proportions of the two error types. After repairing the defects, we score the same rollouts from released policies with both checkers. The original and repaired checkers disagree on up to 9.1\% of a policy's rollouts.

\section{Related Work}
\label{sec:related}

\paragraph{Auditing manipulation benchmarks.}
Recent audits of simulated manipulation benchmarks~\cite{james2019rlbenchrobotlearningbenchmark,yu2021metaworldbenchmarkevaluationmultitask,mees2022calvinbenchmarklanguageconditionedpolicy,tao2025maniskill3gpuparallelizedrobotics,pumacay2024colosseumbenchmarkevaluatinggeneralization,zhang2024vlabenchlargescalebenchmarklanguageconditioned,nasiriany2024robocasalargescalesimulationeveryday,li2024behavior1khumancenteredembodiedai,lu2026planning} examine two things. The first is the task. LIBERO-PRO~\cite{zhou2026liberoprorobustfairevaluation} and LIBERO-Plus~\cite{fei2025liberoplusindepthrobustnessanalysis} perturb objects, layouts and instructions, and Jiang \etal~\cite{jiang2026actuallybenchmarkingrobotmanipulation} test five suites for shortcut solvability, statistical significance, overfitting and data-source dependence. The second is the simulator~\cite{li2024evaluatingrealworldrobotmanipulation}. SimuGuard~\cite{zhou2026simulatorfailspolicyauditing} traces contact artifacts in RoboTwin to one constant in the physics engine and shows that success on container tasks rises in a repaired engine. SimuGuard also notes in passing that RoboTwin's checks read the commanded gripper value, and we measure what that choice does to verdicts. 
These audits use the benchmarks' own success checkers to measure how changes to tasks or simulator physics affect reported success rates.
This approach assumes that the checker verdicts accurately reflect whether the task was completed.
We test that assumption directly by holding the task and simulator fixed and checking whether the verdicts
agree with the task instructions.

\paragraph{Auditing benchmarks for software agents.}
Auditing how benchmarks~\cite{zhou2024webarenarealisticwebenvironment,xie2024osworldbenchmarkingmultimodalagents,yao2024taubenchbenchmarktoolagentuserinteraction} score agent outputs is an established practice in software-agent evaluation~\cite{liu2023codegeneratedchatgptreally,aleithan2024swebenchenhancedcodingbenchmark}. 
The Agentic Benchmark Checklist~\cite{zhu2025establishingbestpracticesbuilding} identifies \emph{outcome validity}, the extent to which passing an evaluation check indicates that the task was completed, as a core requirement.
UTBoost~\cite{yu2025utboostrigorousevaluationcoding} augments the test suites in SWE-bench~\cite{jimenez2024swebenchlanguagemodelsresolve} and changes the scores of 24--41\% of leaderboard entries. 
SWE-bench Verified~\cite{openai2024swebenchverified} improves evaluation reliability through task curation, and recent work extends automated auditing to a broader range of benchmarks~\cite{wang2026automatedbenchmarkauditingai,wang2026androidsdreambreakinggame}.
Software-agent audits can use artifacts such as execution logs and test results to assess disputed outcomes. 
In simulated manipulation, testing a suspected checker defect may first require constructing an execution that exposes it. 
For example, determining whether a robot shook a bottle requires evidence of its motion over time. 
\simaudit{} constructs targeted cases in simulation, uses physics-based robot motion where possible, and compares the original checker's verdict with evidence from multi-view renderings. 

\paragraph{Success checkers in manipulation.}
Some work replaces the hand-written binary checker~\cite{du2023visionlanguagemodelssuccessdetectors,rocamonde2024visionlanguagemodelszeroshotreward,liu2023reflectsummarizingrobotexperiences,duan2024ahavisionlanguagemodeldetectingreasoning}. RoboEval~\cite{wang2026roboevalroboticmanipulationmeets} adds process metrics, and AutoEval~\cite{DBLP:journals/corr/abs-2503-24278} uses learned judges, which have errors and biases of their own~\cite{huang2026mllmjudges,xu2026unfairjudge}: the best VLM-based detector in FailBench reaches 0.77 balanced accuracy~\cite{navasardyan2026failbenchreliablevlmsjudging}. 
Leaderboards still count the verdicts of hand-written or LLM-written checkers~\cite{wang2024robogenunleashinginfinitedata,wang2024gensimgeneratingroboticsimulation}, and a survey of verifiers for robot policies~\cite{wan2026freecheckersurveyverifiers} observes that their error rate is reported far less often than the success rate they produce. 
The study most closely related to ours is a manual audit of LIBERO and its variants, as well as RoboTwin, RoboCasa, and VLABench~\cite{chen2026fasterbetterbenchmarkbugs}. The audit identifies 22 bugs and four design limitations. Examples include unrealistic object masses, tasks that can pass without a grasp, and success checkers that impose requirements beyond the task instructions, such as a specific bottle orientation in the ``adjust bottle'' task.
We differ from it in three ways. 
First, our automated search covers every task across three benchmarks, including RoboDojo, which was outside the scope of the prior audit. 
Second, each defect is documented with rendered evidence and can be reproduced by running the unmodified checker on the same simulated state.
Third, we quantify the direction of checker errors in each benchmark and measure how many verdicts for released policies change after the checkers are repaired.

\begin{figure*}[t]
  \centering
  \includegraphics[width=\linewidth]{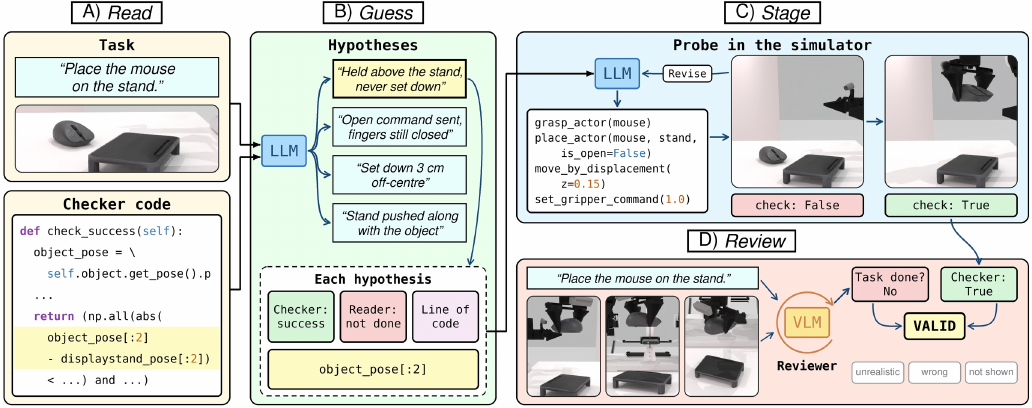}
\caption{\textbf{SimAudit workflow.} \textbf{(A)} An LLM agent reads the task instruction and checker code.
\textbf{(B)} It proposes cases likely to expose checker errors.
\textbf{(C)} A probe in the unmodified simulator triggers a success verdict
while the mouse is still held above the stand.
\textbf{(D)} A separate VLM reviewer uses multi-view renderings to confirm
that the task is incomplete, identifying a \ps{}.}
  \label{fig:pipeline}
\end{figure*}

\section{Auditing Success Checkers}
\label{sec:method}

\subsection{Phantom success and silent failure}
\label{sec:def}
A task comes with an instruction and a checker, a function of the simulator state that returns success or failure. The benchmark ends the episode at the first state on which the checker returns success and records a success; if none occurs before the time limit, it records a failure. A \emph{\ps{}} is credit for an execution that does not fulfil the instruction: the robot does part of the task, does something else, or only passes through the accepted state. A \emph{\sfail{}} is credit withheld from an execution that fulfils it: the robot completes the task by a route, or to an end state, that the instruction allows and the checker does not. A person who reads the instruction and looks at the scene decides whether it is fulfilled. Both are judged on the state at which the benchmark gives its verdict; what the robot would have done after the episode ends does not count.

A \emph{checker defect} is a flaw in a checker together with a state that shows it. We count a defect when three conditions hold. (i)~The state is rendered, and the disagreement between the checker and the person is clear from the pictures. (ii)~The checker value is read from the unmodified benchmark code at that state. (iii)~The state is reachable: it is produced by motions of the robot or of the scripted expert under physics, or it is a pose in which an object rests after it is released and left to settle. A state that exists only while an object is teleported into place, such as an object floating without support or two objects interpenetrating, shows what the code computes but is not counted. Each defect is counted once: several states that show the same flaw in the same task are one defect.

\subsection{\simaudit{}: read, guess, stage, review}
\label{sec:pipeline}

\simaudit{} is an automated pipeline that audits a benchmark's success checkers against its task instructions. Given a task, it constructs targeted test executions, records the original checker's verdicts, and uses rendered evidence to verify potential mismatches. We illustrate the pipeline in \cref{fig:pipeline}. The framework could be divided into four stage as following.

\paragraph{Read.} \simaudit{} starts by establishing what the task asks for and what the checker tests. An LLM agent receives the instruction templates given to the policy, the source of the task with its success checker and every helper it calls, the benchmark's evaluation loop, and the scripted expert when the benchmark provides one. The agent first writes, from the instruction alone, what has to be true of the scene for the task to count as done. It then writes, from the code alone, what the checker accepts. We ask for the two descriptions separately so that the agent's reading of the task is not taken from the checker's conditions. As shown in Figure~\ref{fig:pipeline}A, the instruction of ``place object stand'' asks for the object to be placed on the stand, and the checker accepts any state in which the object is within 3\,cm of the stand horizontally and both gripper commands are open; it tests neither the height of the object nor its contact with the stand.

\paragraph{Guess.} Given the two descriptions, the agent proposes up to four hypotheses for the task. Each hypothesis names a state, the verdict the checker will give on it, and the opposite judgement of a person who reads the instruction, together with the line of code responsible. The agent also states how the state can be built and which measurement would confirm the hypothesis, so that the next stage has a prediction to test. As shown in Figure~\ref{fig:pipeline}B, the agent writes four hypotheses for ``place object stand''. The object is held above the stand and never set down, the open command is sent while the fingers are still closed, the object is set down off-centre but within the tolerance, or the stand is pushed along with the object.

\paragraph{Stage.} A hypothesis is a claim about code until the state exists. For each hypothesis the agent writes a probe, a short program against a small API that can run the scripted expert or individual robot motions under physics, set a gripper command, move an object and let it settle, step the simulator, and read the checker. The API exposes the benchmark's own motions so that, wherever possible, a state is reached the way a policy would reach it. The probe runs in the unmodified simulator. The harness records the checker value and the first step at which it is true. The agent then reads the measurements; if the probe failed or did not produce the intended state, the agent revises it. A hypothesis whose probe returns the predicted checker value is \emph{agent-confirmed}. As shown in Figure~\ref{fig:pipeline}C, the probe for the first of these hypotheses grasps a mouse with the benchmark's own motions, carries it over the stand, lifts it and commands the gripper open. The checker is false before the command and true on the next physics step, and the episode ends with the mouse still in the gripper, 21\,cm above the stand.

\paragraph{Review.} An agent-confirmed hypothesis rests on a value returned by a probe, and the probe was written by the agent whose hypothesis it tests. The value does not show what the scene looks like: a probe can place an object below the table top, or describe a bottle as lying down when the asset's axis makes it stand. \simaudit{} therefore re-runs every agent-confirmed probe with rendering. Cameras aimed at the manipulated objects capture the scene after every teleport, at the first success step, again after a short wait to make sure the accepted state persists, whenever the checker value flips, and at the end. A reviewer that did not write the probe sees the instruction, the frames and the checker value at each frame, and never the code, so that its judgement of the task cannot follow the checker's logic. It assigns one of four labels. \emph{valid}: the frames show the disagreement that the hypothesis claims, on a reachable state. \emph{unrealistic}: the checker behaves as claimed, but the state exists only by teleportation. \emph{wrong}: the frames contradict the claim, or the state the checker accepts is a reasonable success. \emph{not shown}: the frames do not show the claimed state. Only \emph{valid} items are counted. As shown in Figure~\ref{fig:pipeline}D, the reviewer VLM sees a mouse held in the air and a checker value of success, answers that the task is not done, and labels the item \emph{valid}.

\begin{figure*}[p]
  \centering
  \includegraphics[width=\linewidth]{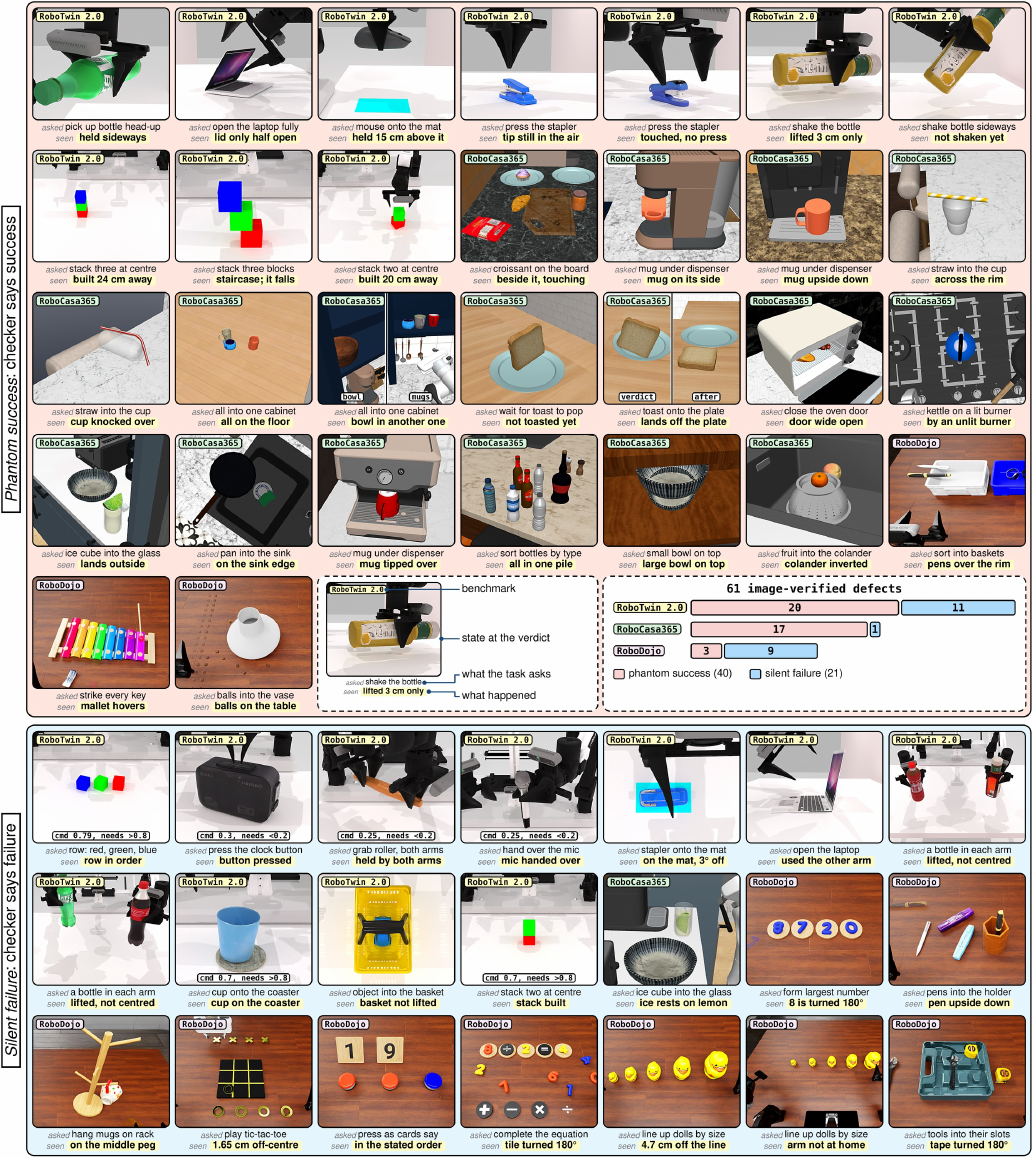}
\caption{\textbf{Verified checker defects across three benchmarks.}
Each tile pairs the task instruction with the observed state at the checker’s verdict. Phantom successes appear above and silent failures below. \textit{cmd} denotes the commanded gripper value. The tiles show 51 defects, and the inset counts all 61, including ten additional RoboTwin~2.0 successes reported while the object was still held.}

  \label{fig:defects}
\end{figure*}

\subsection{From defects to Verified checkers}
\label{sec:verified}
A defect confirmed by \simaudit{} comes with the state that shows it and the line of code responsible. This is what a repair needs. We repair each checker so that its verdict on that state agrees with the instruction. The tasks, assets and episode protocol stay as released. Thresholds stay as released unless a defect concerns the threshold itself. We call the result a \emph{Verified} checker.

A checker reaches its verdict in three steps. It measures quantities of the scene, it tests conditions on these quantities, and it decides at which moment the verdict is final for the episode. A defect enters at one of these steps. The checker measures the wrong quantity, it tests too few or too many conditions, or it gives its verdict too early. We therefore repair a checker step by step, in the order in which the steps depend on each other.

\paragraph{Measure the physical state.} A condition is only as reliable as the quantity it is tested on, so the repair starts with the measurement. Some simulation benchmarks decide success from a quantity that stands in for the state of the scene. For example, a checker may read the commanded gripper value to decide whether an object has been released. The Verified checker measures the physical state directly. A gripper holds an object when its links exert force on it, and has released it when they do not. For example, the original checker accepts the mouse held above the stand as soon as the open command is sent. The Verified checker rejects this state, because the fingers still press on the mouse.

\paragraph{Match the conditions to the instruction.} Correct measurements are not enough when the conditions tested on them differ from the instruction. The difference has two directions, and each produces one of the two errors. A checker that tests less than the instruction states gives \ps{}es, so the Verified checker adds the missing condition. For example, a bottle must be shaken to count as shaken. A checker that tests more than the instruction states gives \sfail{}s, so the Verified checker removes every condition that the instruction does not mention. For example, an instruction that only asks the robot to pick up an object does not require the object to end at a fixed target position.

\paragraph{Require success to persist.} A checker with correct measurements and correct conditions can still give its verdict too early. The conditions can hold for a few steps while an object falls or a stack topples, and the episode ends at the first of these steps. The Verified checker therefore counts a success only after the repaired conditions have held for one second. For example, three blocks are stacked with an overhang, and the original checker accepts the stack while it is already toppling. The Verified checker rejects this state, because the stack does not stay up for one second.

\paragraph{Validation.} A repair can go wrong in two ways, and we test for both. First, it can make a task harder than the instruction requires. The scripted expert completes every task as the benchmark intends, so it must still pass after repair. The Verified checker accepts 1{,}233 expert episodes. Second, a repair can leave the defect open. The rendered states that revealed the 61 defects therefore serve as a regression suite. In all of these tests the Verified checker runs as an observer next to the original. It reads the simulator state at every step at which the original checker is evaluated and never changes it.
\section{Benchmark Audit Results}
\label{sec:exp}

\subsection{Audit Setup}
We audit three benchmarks: RoboTwin~2.0~\cite{DBLP:journals/corr/abs-2506-18088}, RoboCasa365~\cite{DBLP:conf/iclr/NasirianyNMZ26} and RoboDojo~\cite{chen2026robodojounifiedsimandrealbenchmark}. All checker values come from unmodified benchmark code. We assign each defect one primary cause. A checker can omit a condition that the instruction requires, or add a condition that the instruction does not state. It can read an indirect signal instead of the physical state, or accept a region that is too large. It can also give its verdict before the scene settles. We call these five causes \emph{missing condition}, \emph{unstated requirement}, \emph{proxy signal}, \emph{loose geometry} and \emph{transient success}, as detailed in Appendix~\ref{sec:app-taxonomy}.

\subsection{Audit Findings Across Benchmarks}
\label{sec:common}

\begin{wraptable}{r}{0.5\linewidth}
  \vspace{-10pt}
  \centering
  \footnotesize
    \caption{\textbf{Verified checker defects in the three benchmarks.}  The defects are split first by the direction of the error and then by their primary cause.}
  \setlength{\tabcolsep}{5pt}
  \resizebox{\linewidth}{!}{%
  \begin{tabular}{@{}lcccc@{}}
    \toprule
     & RoboTwin~2.0 & RoboCasa365 & RoboDojo & Total \\
    \midrule
    Defects & 31 & 18 & 12 & 61 \\
    Tasks with a defect & 25 & 13 & 11 & 49 \\
    \midrule
    \multicolumn{5}{@{}l}{\emph{By direction of the error}} \\
    \hspace{0.6em}\textcolor{pscol}{\Ps{}} & 20 & 17 & 3 & 40 \\
    \hspace{0.6em}\textcolor{sfcol}{\Sfail{}} & 11 & 1 & 9 & 21 \\
    \midrule
    \multicolumn{5}{@{}l}{\emph{By primary cause}} \\
    \hspace{0.6em}Missing condition & 10 & 7 & 0 & 17 \\
    \hspace{0.6em}Unstated requirement & 5 & 1 & 9 & 15 \\
    \hspace{0.6em}Proxy signal & 15 & 0 & 1 & 16 \\
    \hspace{0.6em}Loose geometry & 0 & 8 & 2 & 10 \\
    \hspace{0.6em}Transient success & 1 & 2 & 0 & 3 \\
    \bottomrule
  \end{tabular}}
  \label{tab:main}
  \vspace{-6pt}
\end{wraptable}
\Cref{tab:main} gives the main result, and Figure~\ref{fig:defects} shows the defects at the state where the original checker gives its verdict. The three benchmarks do not share a failure profile, as detailed in Appendix~\ref{sec:app-bench}.

RoboTwin~2.0 errs in both directions. Its main cause is the proxy signal, as detailed in Appendix~\ref{sec:gripper}. The checkers read the commanded gripper value instead of the state of the fingers. A placement is credited when the open command is issued with the object still in the gripper, and a finished grasp is rejected when the command stops short of a threshold. RoboCasa365 over-credits. Most of its defects are loose geometry or missing conditions. For example, a straw lying across the rim of a cup counts as inside the cup. RoboDojo under-credits. Most of its defects are silent failures, and all of these are unstated requirements. For example, the digit 8 is rejected after a half turn that the instruction does not mention and the picture does not show. The statement that simulated success rates are inflated therefore holds for RoboCasa365, holds in part for RoboTwin~2.0, and is reversed for RoboDojo.
The cause of a defect fixes its direction. Missing conditions, loose geometry and transient successes are all phantom successes, and all unstated requirements are silent failures. The proxy signal is the one cause that errs in both directions, with 10 phantom successes and 6 silent failures.

\begin{figure*}[t]
  \centering
  \includegraphics[width=\linewidth]{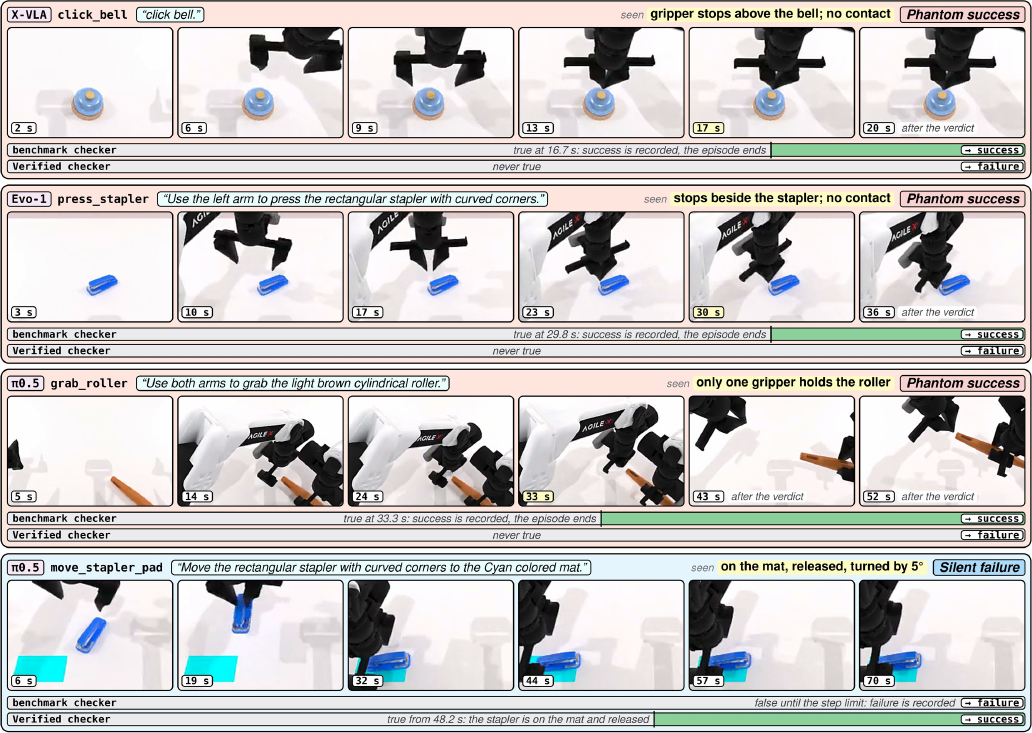}
\caption{\textbf{Checker verdicts on four RoboTwin~2.0 rollouts.} Each row shows a rollout scored by the original and Verified checkers; green denotes a true verdict. The top three rows show \ps{}es: the gripper stops above the bell or beside the stapler, or holds the roller with one gripper. The bottom row shows a \sfail{}: the stapler is placed on the mat, yet the original checker rejects it for a yaw requirement absent from the instruction.}

  \label{fig:rollouts}
\end{figure*}

\subsection{Evaluation of \simaudit{}}
\label{sec:evalsimaudit}

We evaluate \simaudit{} by asking independent annotators to verify the reported defects and examining which candidates are filtered out during visual review.

\paragraph{Precision.} Two annotators independently examine each reported defect. They see the instruction, the rendered frames and the checker's verdict, and they decide whether the verdict is wrong. Both annotators confirm every defect.

\paragraph{Effect of the review stage.}
Of the 183 candidates for which the checker returned the predicted verdict, approximately two-thirds were rejected during visual review. Appendix~\ref{sec:app-eval} breaks down the reasons for rejection. The annotators then examined a random sample of 50 rejected candidates and found no valid defects among them.

\subsection{Policy Re-evaluation with Verified Checkers}
\label{sec:rerank}
The defects above are states we constructed. We now ask how often released policies reach them, and whether the repair changes how the policies rank. On RoboTwin~2.0 we evaluate five policies with public weights: LingBot-VA~\cite{li2026causalworldmodelingrobot}, LingBot-VLA~2.0~\cite{DBLP:journals/corr/abs-2607-06403}, Evo-1~\cite{lin2025evo1lightweightvisionlanguageactionmodel}, X-VLA~\cite{zheng2025x} and $\pi_{0.5}$~\cite{DBLP:journals/corr/abs-2504-16054}. On RoboCasa365 we evaluate GR00T~N1.5~\cite{nvidia2025gr00tn1openfoundation} fine-tuned on 100\% and on 30\% of the training data, Xiaomi-Robotics-1~\cite{xiaomiroboticsteam2026xiaomirobotics1scalingvisionlanguageactionmodels} and REKI-2B~\cite{marrie2026rekiregisteringknowledgevlas} fine-tuned on RoboCasa365, and the $\pi_{0.5}$ and $\pi_0$~\cite{black2026pi0visionlanguageactionflowmodel} checkpoints released with the benchmark. On RoboDojo we evaluate Rex-M1, VPP~\cite{hu2025videopredictionpolicygeneralist}, Awomo-0.5~\cite{team2026awomo}, Xiaomi-Robotics-1 and two StarVLA~\cite{community2026starvlalegolikecodebasevisionlanguageaction} models with different action heads QwenPI and QwenOFT. Each policy runs with its released inference code under the benchmark's official protocol, as detailed in Appendix~\ref{sec:app-retest}.

We score each rollout with both the original and Verified checkers. The original checker records an episode as successful at its first success verdict. During re-evaluation, policy execution continues beyond this point, and the Verified checker can reach success on subsequent states of the same rollout. An episode accepted by the original checker but rejected by the Verified checker is counted as a phantom success; the reverse is counted as a silent failure.

\begin{table}[t]
  \centering
  \setlength{\tabcolsep}{3pt}
  \caption{\textbf{Re-scoring released policies on the same rollouts.}
$\Delta$ is the Verified minus original success rate in percentage points.
\emph{Phan.\ succ.} and \emph{Silent fail.} are the percentages of original
successes and failures reversed, respectively. \emph{Flip} is the percentage
of all episodes whose verdict changes.}
  \IfFileExists{tab/retest.tex}{\begin{tabular*}{\linewidth}{@{\extracolsep{\fill}}lrrrrrr@{}}
\toprule
 & \multicolumn{3}{c}{Success (\%)} & \textcolor{pscol}{Phan.} & \textcolor{sfcol}{Silent} & \\
\cmidrule(lr){2-4}
Policy & Orig. & Verif. & $\Delta$ & \textcolor{pscol}{succ.} & \textcolor{sfcol}{fail.} & Flip \\
\midrule
\multicolumn{7}{@{}l}{\emph{RoboTwin~2.0}} \\
LingBot-VA & 93.1 & 90.0 & -3.1 & 3.3 & 0.0 & 3.1 \\
LingBot-VLA 2.0 & 93.9 & 90.8 & -3.1 & 3.4 & 1.5 & 3.3 \\
Evo-1 & 68.3 & 65.5 & -2.7 & 5.8 & 3.9 & 5.2 \\
X-VLA & 64.9 & 61.6 & -3.4 & 6.2 & 2.0 & 4.7 \\
$\pi_{0.5}$ & 46.2 & 42.3 & -3.9 & 10.2 & 1.6 & 5.6 \\
\midrule
\multicolumn{7}{@{}l}{\emph{RoboCasa365}} \\
GR00T~N1.5 (100\%) & 41.2 & 38.8 & -2.4 & 5.8 & 0.0 & 2.4 \\
GR00T~N1.5 (30\%) & 37.7 & 33.8 & -3.8 & 12.2 & 1.2 & 5.4 \\
Xiaomi-Robotics-1 & 38.2 & 34.8 & -3.4 & 12.5 & 2.2 & 6.1 \\
REKI-2B & 29.8 & 28.2 & -1.5 & 5.1 & 0.0 & 1.5 \\
\midrule
\multicolumn{7}{@{}l}{\emph{RoboDojo}} \\
Rex-M1 & 35.5 & 40.0 & +4.5 & 0.0 & 7.0 & 4.5 \\
VPP & 30.0 & 39.1 & +9.1 & 0.0 & 13.0 & 9.1 \\
Awomo-0.5 & 20.0 & 21.0 & +1.0 & 0.0 & 1.2 & 1.0 \\
Xiaomi-Robotics-1 & 10.9 & 11.8 & +0.9 & 0.0 & 1.0 & 0.9 \\
StarVLA (QwenPI) & 5.5 & 9.1 & +3.6 & 0.0 & 3.8 & 3.6 \\
StarVLA (QwenOFT) & 0.9 & 5.5 & +4.5 & 0.0 & 4.6 & 4.5 \\
\bottomrule
\end{tabular*}
}{\tbd{tab/retest.tex is generated by scripts/make\_retest\_table.py}}

  \label{tab:retest}
\end{table}

\begin{figure*}[t]
  \centering
  \includegraphics[width=\linewidth]{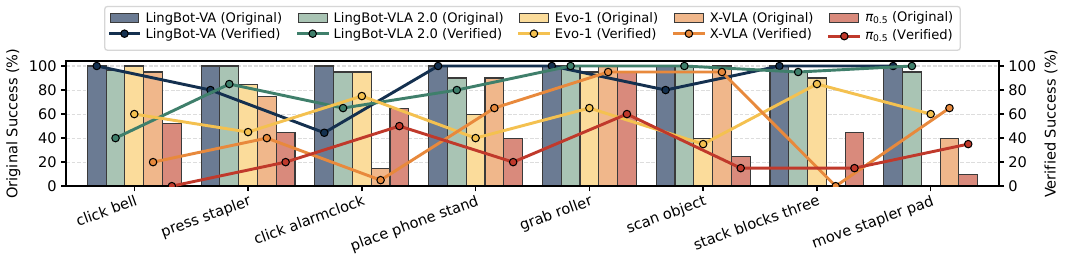}
\caption{\textbf{Per-task success rates.} Bars show the original checker and lines the Verified checker on the same rollouts.}
  \label{fig:retesttasks}
\end{figure*}

\paragraph{Before and after the repair.} Figure~\ref{fig:rollouts} compares the two checkers on four RoboTwin~2.0 rollouts. Each row shows the frames of one episode, and the bars under the frames show the value of each checker over the time axis. In the first row, X-VLA lowers the gripper towards the bell and stops above it. The original checker turns true, so the benchmark records a success and ends the episode. The Verified checker is never true, because the gripper never touches the bell. The second and third rows follow the same pattern. Evo-1 stops beside the stapler without touching it, and $\pi_{0.5}$ holds the roller with one gripper although the instruction asks for both. In the fourth row the comparison is reversed. $\pi_{0.5}$ places the stapler on the mat and releases it. The original checker stays false until the limit because it is turned by 5$^\circ$, so the benchmark records a failure. 
The Verified checker turns true as soon as the stapler rests on the mat.

\paragraph{Verdicts change on every benchmark.}
Rescoring the same rollouts with the Verified checker changed episode verdicts across all three benchmarks and for nearly every policy. The share of episodes affected varied by policy, with the largest change observed for VPP on RoboDojo. Verdicts remained unchanged only for $\pi_{0.5}$ and $\pi_0$ on RoboCasa365. As \Cref{tab:retest} shows, the share of episodes with changed verdicts can exceed the net change in success rate because some episodes gain credit while others lose it.

\paragraph{The direction follows the audit.} The sign of the change on each benchmark matches the audit: success rates fall on RoboCasa365, rise on RoboDojo, and drop on RoboTwin~2.0 by less than the share of flipped verdicts, because the two errors cancel in part. As shown in Figure~\ref{fig:retesttasks}, the two errors occur in different tasks. Seven of the eight tasks lose successes under the Verified checker, and only ``move stapler pad'' gains them. On ``click bell'', $\pi_{0.5}$ keeps none of the successes it is credited with, because the gripper stops above the bell and never touches it. The same defect does not affect all policies equally. On this task, LingBot-VA keeps all of its successes, and X-VLA falls from 95\% to 20\%. On ``move stapler pad'', Evo-1 is credited with no success and succeeds in most episodes under the Verified checker, because the stapler is on the pad a few degrees off the yaw that the checker demands. A leaderboard that averages over tasks reports neither.

\paragraph{Checker repairs change the gaps between policies.}
Changes in episode verdicts were concentrated in a few tasks, so the effect of rescoring varied with how each policy performed on those tasks. On RoboTwin~2.0, the gaps between policies shifted only slightly. On RoboCasa365, both GR00T~N1.5 variants scored lower under the Verified checker, with a larger drop for the variant trained on less data. The gap between the two variants therefore widened. On RoboDojo, VPP gained more from rescoring than either Rex-M1 or StarVLA (QwenPI), leaving it close to the leading policy and further ahead of StarVLA (QwenPI). As shown in \Cref{tab:retest}, the rankings stayed the same on all three benchmarks, even as rescoring changed how closely some policies matched their nearest competitors.
\section{Conclusion}
\label{sec:conclusion}

Prior audits of simulated manipulation benchmarks have focused on task
design and simulator fidelity, leaving the reliability of success
checkers less explored. We introduced \simaudit{} to test these checkers
beyond expert demonstrations using targeted simulations and multi-view
renderings. Our audit uncovered both \ps{}es and \sfail{}s, and
re-scoring released policy rollouts with repaired checkers showed that
these defects affect reported success rates. We release \simaudit{},
the repaired checkers, and the rendered counterexamples to support future  validations.

\section*{LLM Usage Statement}
We used Claude Code for language editing, grammar
correction, and LaTeX formatting of the manuscript,
as well as for drafting code snippets and debugging scripts.

\bibliography{main,extra}
\clearpage
\clearpage
\appendix

\section{Defect taxonomy}
\label{sec:app-taxonomy}

\begin{table}[t]
  \centering
  \small
  \setlength{\tabcolsep}{4pt}
  \begin{tabular}{@{}lp{4.95cm}@{}}
    \toprule
    Cause & The checker \ldots \\
    \midrule
    \multicolumn{2}{@{}l}{\emph{Specification: tests conditions that differ from the instruction}} \\
    \quad Missing condition & omits a required condition, such as bottle motion in ``shake bottle''. \\
    \quad Unstated requirement & adds a condition, such as a centre target in ``pick dual bottles''. \\
    \midrule
    \multicolumn{2}{@{}l}{\emph{Measurement: tests a condition on an unsuitable quantity}} \\
    \quad Proxy signal & reads an indirect signal, such as the commanded gripper value for a grasp. \\
    \quad Loose geometry & uses a region or threshold that is too permissive, such as a straw across the rim of a cup counted as inside it. \\
    \midrule
    \multicolumn{2}{@{}l}{\emph{Timing: gives its verdict too early}} \\
    \quad Transient success & accepts a condition that holds briefly while an object falls, slides or topples. \\
    \bottomrule
  \end{tabular}
  \caption{\textbf{Checker defect taxonomy.} Five causes in three categories. Each category corresponds to one step of a checker's verdict.}
  \label{tab:taxonomy}
\end{table}

\Cref{tab:taxonomy} lists the five causes, grouped by the step of the verdict at which the checker diverges from the task instruction.

\section{Defects of each benchmark}
\label{sec:app-bench}
\textit{RoboTwin~2.0.} The phantom successes come from missing conditions (10), the gripper proxy (9) and a transient success (1). The missing conditions are a shake that ``shake bottle'' never tests, an orientation that ``adjust bottle'' never tests, and a location for a stack that the instruction places at the centre. The silent failures come from the same proxy (6) and from unstated requirements (5).

\textit{RoboCasa365.} Many defects trace to shared placement helpers that test contact and horizontal distance only, and every task that calls a helper inherits its tolerance. Fruit beside an overturned colander is in the colander; a mug upside down under the spout is ready for coffee.

\textit{RoboDojo.} The most frequent unstated requirement is the orientation of an object that looks the same after a half turn: the digit 8, an operator tile, a tape measure.

\section{Gripper commands and early termination}
\label{sec:gripper}
RoboTwin's checkers decide whether a gripper is open by reading the last commanded value, and require it to be above 0.8 for a release or below 0.2 for a grasp~\cite{zhou2026simulatorfailspolicyauditing}. The command is not the state of the fingers, and the proxy fails in both directions: 9 phantom successes and 6 silent failures.

The silent failures arise because the thresholds are strict. In ``place empty cup'' the cup stands on its coaster and the arm has left, and the checker returns failure for as long as the command is 0.7. A policy that outputs continuous gripper actions and stops short of the extremes completes the task and is scored as failing.

The phantom successes arise because a command is issued before the fingers move, and because the episode ends too early to see them move. The checker turns true on the first physics step after the open command, with the fingers still closed on the object. The episode ends at that step, so this momentary agreement becomes the final verdict. In ``place mouse pad'' the mouse is recorded as placed while it is held above the mat, and in ``stamp seal'' the seal has not touched the mat.

Ten RoboTwin defects are successes recorded while the object is still in the gripper. Eight come from the gripper proxy. The other two are in ``place bread skillet'' and ``put bottles dustbin'', where the checker has no release condition at all. To see what the recorded episode leaves out, we continued these ten probes past the first success step and kept the open command. In all ten the object ended in the correct place: it dropped a few centimetres into its slot, onto its plate or onto its mat, or stayed in the receiving hand. A policy that goes on to release is therefore credited early, and a policy that stops there is credited for a task it did not finish. The recorded episode is the same in both cases.

Early termination also hides the opposite case, in which the state is right at the first success step and wrong afterwards. In ``stack blocks three'', three blocks are placed with an overhang that is within tolerance for each pair and accumulates across the stack. They pass the checker at the first step. Within 60\,ms, the stack is falling and the checker value is false again, but the episode has already been recorded as a success.

\section{Evaluation of \simaudit{}: details}
\label{sec:app-eval}
\paragraph{Rejected states by label.} On RoboTwin~2.0, the 91 agent-confirmed hypotheses split into 31 distinct defects, 1 repeat of one of them, 37 \emph{unrealistic}, 19 \emph{wrong} and 3 \emph{not shown}.

\paragraph{Cost.} Auditing one task costs \$0.78 in LLM calls and takes 18 minutes on average over the 142 tasks we audited, including the time to run the probes in the simulator.

\section{Policy re-evaluation details}
\label{sec:app-retest}
\paragraph{Protocol.} On RoboTwin~2.0, each policy runs with its released inference code on all 50 tasks of the clean setting, with the official seeds, expert filter and unseen instructions. For $\pi_{0.5}$ we use the checkpoint released with Motus~\cite{bi2025motusunifiedlatentaction}.

\begin{table}[t]
  \centering
  \small
  \setlength{\tabcolsep}{3pt}
  \setlength{\tabcolsep}{2.5pt}\IfFileExists{tab/retest_cause.tex}{\begin{tabular}{@{}lrrrrr@{}}
\toprule
 & \multicolumn{2}{c}{LingBot} & & & \\
\cmidrule(lr){2-3}
Verdict change & VA & VLA~2 & Evo-1 & X-VLA & $\pi_{0.5}$ \\
\midrule
Press without contact & 6 & 23 & 20 & 24 & 19 \\
Missing condition & 0 & 2 & 4 & 3 & 9 \\
Command-based grasp & 0 & 0 & 5 & 3 & 7 \\
Transient success & 2 & 9 & 13 & 11 & 15 \\
\midrule
Silent failures & 0 & 1 & 13 & 7 & 9 \\
\bottomrule
\end{tabular}
}{\tbd{tab/retest\_cause.tex}}
\caption{\textbf{Verdict changes by repair type and policy.}
The first four rows count corrected \ps{}es, and the fifth counts
corrected \sfail{}s.}
  \label{tab:retestcause}
\end{table}

\paragraph{Protocol on RoboCasa365 and RoboDojo.} On RoboCasa365 we evaluate 13 tasks in which the audit found a defect. Every episode runs to the task horizon. On RoboDojo we use the released policy servers and evaluation client, and evaluate 11 tasks in which the audit found a defect, on layouts 0 to 9 of each task. On both benchmarks the two checkers score the same rollout, as on RoboTwin~2.0.

\paragraph{A few repairs account for most flips.} \Cref{tab:retestcause} attributes each flipped verdict to the repair that caused it. The largest group is pressing without touching: in ``click bell'', ``click alarmclock'' and ``press stapler'' the original checker accepts a contact point that the simulator reports across an air gap of up to 2\,cm. The second group is success that does not last: a phone that tips over within a second of being placed on its stand. The flips concentrate in 4 to 19 of the 50 tasks, and they are largely the same tasks for every policy.

\end{document}